\documentclass{article}

\usepackage{PRIMEarxiv}

\usepackage[utf8]{inputenc} 
\usepackage[T1]{fontenc}    
\usepackage{hyperref}       
\usepackage{url}            
\usepackage{booktabs}       
\usepackage{amsfonts}       
\usepackage{nicefrac}       
\usepackage{microtype}      
\usepackage{lipsum}
\usepackage{fancyhdr}       
\usepackage{graphicx}       
\graphicspath{{media/}}     
\usepackage{tcolorbox}
\usepackage{multirow}
\usepackage{comment}

\title{Error-Aware Knowledge Distillation via Targeted Revision for Customer-Service Summarization
}

\author{
  Hee-Jin Lee, Zhen Guo, 
  Luchao Jin,
  Morteza Moazami Goudarzi\textsuperscript{*}\\\
  eBay Inc. 2025 Hamilton Avenue, San Jose, CA, USA\\
  \mbox{*} Corresponding author: Morteza Moazami Goudarzi (\texttt{mmoazamigoudarzi@ebay.com}) \\
}

\begin{document}
\maketitle

\begin{abstract}
We introduce an Analyze–Revise–Finetune (ARF) pipeline that enables smaller open-source language models (LLMs) to surpass substantially larger proprietary models in customer service summarization tasks. The pipeline first analyzes and categorizes common errors in summaries produced by a teacher model (GPT-3.5), then performs a targeted revision using a compact editor model (Llama 3.1 70B) to generate high-quality, refined training data. Fine-tuning smaller student models (e.g., Llama 3.1 8B, QWen3 4B) on this refined data resulted in superior summarization performance compared to GPT-3.5. The ARF pipeline improves cost efficiency and data privacy while maintaining competitive accuracy, illustrating a generalizable framework for enhancing open-source LLMs across diverse downstream applications.
\end{abstract}

\keywords{LLM fine-tuning \and Knowledge distillation \and Summarization}

\section{Introduction}

We are in a new era of large language models (LLMs), with numerous products widely deploying these powerful models. This widespread deployment necessitates continuous improvement and monitoring of model quality, particularly concerning factors such as output quality, privacy, and cost. Proprietary LLMs, like GPT, are commonly employed in production environments due to their high-quality outputs achievable through prompt engineering. However, reliance on such large proprietary models carries significant drawbacks, including high costs and the potential risk of price increases. Additionally, off-site hosting of these models introduces privacy and data security concerns, as sensitive data must be transmitted to external servers. Consequently, utilizing smaller, open-source LLMs tailored through task-specific fine-tuning is becoming increasingly preferable, offering reduced inference costs and mitigating privacy risks through on-premises deployment.

Task-specific fine-tuning substantially improves the quality of text generation in smaller LLMs. However, creating a sufficiently large dataset for training can become a critical bottleneck, especially when such data must be generated manually. To address this issue, recent methodologies have utilized pseudo-labeling strategies, where text generated by larger (teacher) LLMs serves as training labels for smaller (student) LLMs~\cite{hinton2015distillingknowledgeneuralnetwork,gu2024minillmknowledgedistillationlarge,sanh2020distilbertdistilledversionbert}. These approaches enable student models to achieve performance close to that of teacher models. However, they also inadvertently propagate any errors present in teacher-generated texts, as all outputs from the teacher model are accepted as correct.

To enhance the quality of pseudo-labels produced by teacher models, recent research has explored self-improvement methods for text generation~\cite{madaan2023selfrefineiterativerefinementselffeedback,peng2023checkfactstryagain,bai2022constitutionalaiharmlessnessai}. These methods typically involve iterative critic-and-revision cycles, in which an LLM evaluates its own output and subsequently revises the text based on its self-critique. Although this iterative approach results in improved final outputs, it incurs substantial computational and inference costs, particularly when applied across extensive training datasets. Moreover, the critic and generation tasks typically rely on the same advanced LLM, increasing the overall inference expense.

\begin{figure*}[t] 
    \centering
    \includegraphics[width=0.95\textwidth]{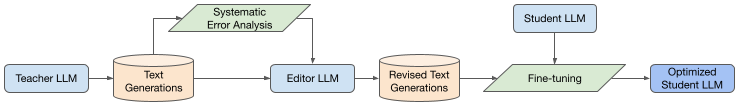} 
    \caption{"Analyze-Revise-Finetune" pipeline}
    \label{fig:overall_process}
\end{figure*}

In this paper, we introduce an \emph{Analyze-Revise-Finetune} (ARF) pipeline (illustrated in Figure~\ref{fig:overall_process}) designed to enhance the performance of smaller, open-source LLMs (student models) so they surpass the capabilities of larger, proprietary teacher models. ARF comprises three primary steps: 1) systematic identification of prevalent error patterns in auto-generated texts, 2) targeted error correction using a separate editor LLM, and 3) leveraging the corrected texts to fine-tune a smaller open-source student LLM. Unlike previous studies that either exclusively depend on teacher-generated pseudo-labels or iterative self-refinement, ARF incorporates a targeted error correction step based on an error-focused evaluation scheme that facilitates a systematic analysis of teacher-generated texts to pinpoint prevalent errors in the training data. By specifically targeting these common errors for correction, we remove the need for repetitive critic-and-revision cycles. Furthermore, targeted error correction is inherently simpler than full critic-revision, allowing the editor LLM to be smaller and more cost-effective than the teacher model. By fine-tuning the student LLM using the revised training data, the student model learns effectively from the teacher without inheriting its frequent errors, unlike methods that uncritically adopt all teacher-generated pseudo-labels.

We demonstrate the effectiveness of ARF 
pipeline in a real-world customer service interaction summarization task that was served by GPT-3.5. 
Initially, we define comprehensive evaluation criteria to systematically identify and classify specific error patterns present in GPT-3.5-generated summaries. Utilizing these criteria, we identify common errors prevalent across summaries. Subsequently, we employ an open source editor model (Llama 3.1 70B) to specifically revise these identified errors, thus creating a refined training dataset. Finally, we fine-tune smaller open-source models on this revised dataset, demonstrating that our fine-tuned models successfully replicate GPT-3.5’s functionality while significantly reducing frequent errors, ultimately outperforming GPT-3.5 in summary generation.

\section{Related Work}
%

Many studies leverage larger models (\emph{teacher} models) to generate pseudo-labels for training smaller models (\emph{student} models). 
Hinton et al.~\cite{hinton2015distillingknowledgeneuralnetwork} introduced a knowledge distillation method where the teacher model's output probabilities serve as soft targets for the student model training. This approach enables the student to capture nuanced predictions from the teacher, effectively consolidating knowledge from an ensemble of models into a single model. 
Gu et al.~\cite{gu2024minillmknowledgedistillationlarge} extended knowledge distillation to large language models (LLMs), employing reverse Kullback-Leibler divergence (KLD) instead of the conventional forward KLD, comparing the student LLM's output distribution directly with that of the teacher. 
Sanh et al.~\cite{sanh2020distilbertdistilledversionbert} developed DistilBERT, which achieves 97\% of BERT's language understanding capability while being 40\% smaller by minimizing the KL divergence between student and teacher softmax outputs. 
Unlike these approaches, which typically treat all teacher-generated outputs equally without any error analysis or correction, ARF provides a focused approach where common errors in teacher-generated texts are explicitly identified and corrected, thereby ensuring higher-quality pseudo-labels.

Recent research has investigated iterative refinement techniques involving LLMs, where a model evaluates and improves its own generated outputs. 
\emph{Self-Refine}~\cite{madaan2023selfrefineiterativerefinementselffeedback} is a framework that allows an LLM to iteratively critique and refine its responses autonomously without external supervision or additional training data. 
Similarly, LLM-AUGMENTER~\cite{peng2023checkfactstryagain} addresses hallucinations in LLM outputs by integrating external knowledge and iterative prompt revision based on feedback from utility functions. 
Bai et al.~\cite{bai2022constitutionalaiharmlessnessai} proposed Constitutional AI, a self-supervised framework that utilizes guiding principles (a "constitution") to critique and iteratively revise an LLM's outputs, ultimately using these revised outputs to train a safer model.
In contrast to these iterative self-refinement methods, ARF 
emphasizes targeted correction of the most common errors identified through systematic analysis. Additionally, while previous methods generally rely on a single LLM for both output generation and self-correction, ARF 
separates initial generation (performed by teacher LLM) from subsequent error correction (performed by editor LLM). This separation enables the use of a smaller and more cost-efficient editor model, achieving high-quality revisions with lower inference costs.



\section{Background: Customer Service Interactions Summarization with GPT3.5}
We base our study on a real-world use case using data from a large e-commerce platform. The platform launched an LLM-based summarization system to improve customer service agent productivity. In typical support workflows, agents must understand a customer’s issue by reviewing prior interactions across multiple communication channels. These interactions can span multiple sessions and contain lengthy, unstructured dialogue, making it time-consuming to reconstruct context before resolving the issue.

The system summarizes customer interactions from four channels: bot-customer chat (BotChat), teammate-customer chat (TeammateChat), emails (Email), and webforms (Webform). To enable continuous improvement, it includes a feedback mechanism where teammates provide thumbs-up or thumbs-down ratings. Since its launch in February 2024, the thumbs-down rate has remained below 5\%.

We fine-tune a compact open-source LLM for customer service summarization to replace GPT-3.5 in production, reducing inference costs and mitigating privacy risks associated with proprietary models.

\section{Error-Focused Summary Evaluation Scheme}
\label{sec:eval_criteria}

To improve the production summarizer and support AI governance, we sampled production inputs and summaries and had subject matter experts and the AI governance team manually review them. Together, we developed a structured evaluation scheme to ensure consistent reviews and systematically surface recurring issues, focusing on identifying summary \textit{errors} mapped to predefined error types.

Errors fall into seven categories: \textit{Content}, \textit{Entities}, \textit{Data Elements}, \textit{Customer Type}, \textit{Unnecessary Information}, \textit{Inferred Sentiment}, and \textit{Language}, each with finer-grained sub-labels. \textit{Content} covers missing, incorrect, or fabricated information in general; \textit{Entities} covers missing/incorrect identifiers (e.g., item, order, or customer details); \textit{Data Elements} captures other structured inaccuracies (e.g., dates); \textit{Customer Type} captures buyer/seller misclassification (WebForm only); \textit{Unnecessary Information} flags irrelevant/redundant details; \textit{Inferred Sentiment} flags unsupported emotional inferences; and \textit{Language} covers language identification and translation errors. The full taxonomy is provided in Table~\ref{table:eval_criteria_error_types} (Appendix).

We rate summaries on a 1--5 scale based on error count and severity: 1 = numerous/critical errors; 2 = one or two errors; 3 = minor error without affecting understanding; 4 = acceptable with no distinct errors but potentially not maximally concise; 5 = error-free.

We use this scheme to evaluate summaries from different LLMs and fine-tuned models (Section~\ref{sec:evaluation}) and to systematically aanlyze and quantify common GPT-3.5 error patterns (Section~\ref{sec:systematic_error_analysis}).

\section{LLM Fine-Tuning via Targeted Error-Correction}
\subsection{Data}


We collected 10,000 training, 200 development, and 200 test samples per communication channel.
All personally identifiable information (PII) in customer interaction log were anonymized by substituting them with artificial fake values.

\subsection{Systematic error analysis}
\label{sec:systematic_error_analysis}
A team of subject matter experts audited GPT-3.5 production summaries (100 BotChat, 68 WebForm, 133 Email, 100 TeammateChat) and labeled errors using our predefined taxonomy (Table~\ref{table:eval_criteria_error_types}). We then identified the most frequent error types per channel, summarized in Table~\ref{table:major_error_types}.


\begin{table*}[t]
\centering
\caption{Major error types across communication channels (percentage of total errors per channel).}
\label{tab:major_errors}
\small
\begin{tabular}{lcccc}
\toprule
\textbf{Error Type}
& \textbf{BotChat } 
& \textbf{WebForm} 
& \textbf{Email} 
& \textbf{TeammateChat} \\
\midrule

\multicolumn{5}{l}{\textit{Unnecessary Information}} \\
unn\_content\_requests\_agent& 38.0\% & -- & -- & -- \\
unn\_content\_redundant & -- & 29.2\% & 23.7\% & 9.9\% \\
unn\_content\_courtesy & -- & -- & 13.0\% & 33.0\% \\
unn\_content\_transfer & -- & -- & -- & 15.8\% \\
unn\_content\_webform\_email\_copy & -- & 8.3\% & -- & -- \\
unn\_other & -- & -- & -- & 7.4\% \\

\addlinespace
\multicolumn{5}{l}{\textit{Content Issues}} \\
content\_missing & 16.5\% & -- & -- & -- \\
content\_order & -- & -- & 15.5\% & -- \\
content\_inaccurate & -- & 6.3\% & 6.3\% & 5.4\% \\
nothing\_to\_summarize & 7.6\% & -- & -- & -- \\

\addlinespace
\multicolumn{5}{l}{\textit{Entity / Customer Type}} \\
entity\_missing\_item\_number & -- & 6.3\% & 8.7\% & -- \\
customer\_type\_inaccurate & -- & 14.6\% & -- & -- \\

\addlinespace
\multicolumn{5}{l}{\textit{Inferred Sentiment}} \\
sentiment\_inferred\_frustrated & 24.0\% & -- & -- & -- \\
sentiment\_inferred\_not\_frustrated & 6.3\% & -- & -- & -- \\

\bottomrule
  \label{table:major_error_types}
\end{tabular}
\end{table*}

\subsection{Automatic error-correction}

The target errors for auto-correction were selected based on three criteria: (1) frequency, as determined by the systematic error analysis (Section~\ref{sec:systematic_error_analysis}); (2) manageability, defined as the effectiveness of correcting a given error type using a simple prompting strategy in preliminary experiments; and (3) business criticality, referring to errors whose correction was deemed essential to improve the overall usability and reliability of the generated summaries.

For the BotChat channel, we targeted the error \texttt{unn\_content\_requests\_agent}, defined as unnecessary mentions in summaries that customers requested human assistance, and \texttt{sentiment\_inferred\_frustrated} and \texttt{sentiment\_inferred\_not\_frustrated}, which capture cases where customer frustration (or lack thereof) was inferred despite not being explicitly stated. For WebForm summaries, we selected \texttt{unn\_content\_webform\_email\_copy}, referring to the unnecessary inclusion of customer requests for email copies of submitted webforms. In the TeammateChat channel, we focused on \texttt{unn\_content\_transfer} and \texttt{unn\_content\_courtesy}, denoting the unnecessary inclusion of transfer information and courteous exchanges between the teammate and the customer, respectively. For the Email channel, we evaluated \texttt{unn\_content\_courtesy} in addition to \texttt{content\_order}, indicating misleading or incorrect ordering of information in the summary, and \texttt{unn\_content\_email\_footer}, which reflects unnecessary references to email footer content.
Finally, all the channels, the error \texttt{unn\_content\_redundant}, denoting unnecessary redundant information, was targeted for correction under the assumption that increased conciseness would benefit summaries across all channels.

We processed all 40,000 training samples across the four channels using a cascading sequence of error-correction prompts. Each prompt was designed to correct a specific error type, and the prompts were applied sequentially, with the \texttt{unn\_content\_redundant} correction applied as the final step for all channels. We denote the dataset prior to applying the \texttt{unn\_content\_redundant} correction as revision version 1 (r1), and the dataset after this correction as revision version 2 (r2). The correction prompts corresponding to each error type are shown in Figure~\ref{fig:error_correction_prompts} (Appendix).

Notably, the prompt targeting \texttt{sentiment\_inferred\_frustrated} was intentionally designed to remove all sentiment-related content rather than only inferred frustration. This broader removal does not adversely affect summary quality, as sentiment information is considered non-essential under our evaluation criteria (Section~\ref{sec:eval_criteria}).

We employed Llama 3.1 70B as the editor LLM for performing these error corrections. Being an open-source model, 
Llama 3.1 70B provides reduced privacy risks. 


\subsection{LLM fine-tuning}

We evaluated a suite of open-source LLMs of varying sizes as student models.
Each model was fine-tuned independently using three dataset variants: the original training dataset (org), revision~1 (r1), and revision~2 (r2).
Instruction-tuning samples were constructed using the same summarization prompts as our production GPT-3.5 system.
Fine-tuning was performed with Low-Rank Adaptation (LoRA)~\cite{hu2021loralowrankadaptationlarge}, using $\alpha=16$, $r=8$, and dropout $=0.1$, applied to all linear modules.
To prevent the model from learning to continue the input transcript, we applied prompt loss masking during SFT, computing the language-model loss only over assistant completion tokens (i.e., system and user tokens were excluded from the loss).

To further reduce ambiguity between ``continue the transcript'' and ``produce a summary,'' we prepend a dedicated \texttt{<summary>} sentinel token at the beginning of the assistant completion to explicitly delimit the summary segment and to act as a learned control token for summary-style generation \cite{li2024controltokens,raffel2020t5}.
This choice is motivated by prior evidence that task prefixes / control tokens can reliably steer Transformer outputs across tasks (including summarization) \cite{raffel2020t5,li2024controltokens}, and that special tokens provide an effective mechanism for adapting and controlling pretrained models~\cite{yang2023pasta}.
During inference, we append \texttt{<summary>\textbackslash n<ul><li>} to the prompt to encourage strict adherence to the required output format (an HTML unordered list).
To enforce proper termination and prevent run-on generation, we apply two logit-level constraints during decoding: (i) the EOS token is disallowed until the model emits the closing tag \texttt{</ul>}, and (ii) once \texttt{</ul>} is detected, EOS is forced as the only selectable next token, ensuring the output terminates immediately after a closed HTML list.
Finally, we enabled n‑gram repetition blocking, which prevents the model from generating any previously seen 16‑token span and helps mitigate degenerate repetition loops in the produced HTML bullet summaries.

%

\section{Evaluation}
\label{sec:evaluation}

Summaries produced by the fine-tuned student LLMs were evaluated using our predefined summary evaluation scheme (Section~\ref{sec:eval_criteria}) and assigned ratings on a 1-to-5 scale. Since manual evaluation of summaries can be time-intensive and could potentially slow down experimental workflows, we developed an automatic summary evaluation system leveraging the LLMs-as-a-Judge framework~\cite{liu-etal-2023-g,fu2024gptscore}.

Specifically, we used GPT-4.1 mini and GPT4o as the judge LLMs and performed prompt tuning separately for each communication channel to output ratings on a 1-to-5 scale, based on the original content and corresponding generated summaries. Our automatic evaluation system demonstrated strong correlation with human ratings when benchmarked against other advanced LLM-as-judge methodologies~\cite{liu-etal-2023-g,fu2024gptscore}. 
The Spearman's $\rho$~\cite{zar2005spearman} and endall's $\tau$~\cite{mukaka2012guide} scores for each channel is shown in Table~\ref{tab:llm_as_judge_corr}.

Our primary evaluation metric is the mean score of automatic ratings provided by the auto-evaluators on the test dataset.

\begin{table}
 \caption{LLM-as-judge alignment to human judgement}
  \centering
  \small
  \begin{tabular}{llll}
    \toprule
    \textbf{Channel} & \textbf{$\rho$} & \textbf{$\tau$} & \textbf{Judge LLM} \\
    \midrule
    TeammateChat & 0.5313 & 0.4802 & GPT4.1 mini \\
BotChat & 0.4735 & 0.4225 & GPT4.1 mini \\
WebForm & 0.4612 & 0.4203 & GPT4.1 mini \\
Email & 0.4551 & 0.4405 & GPT4o \\
    \bottomrule
\end{tabular}
\label{tab:llm_as_judge_corr}
\end{table}

\section{Results}
\subsection{ARF improves fine-tuning performance.}

%

We evaluated the effectiveness of the "Analyze-Revise-Finetune" pipeline across various student LLMs. Table~\ref{tab:perf_on_training_set_llm} presents the summarization performance for each model across different training conditions. Each model was assessed under four distinct scenarios: 1) without fine-tuning (out-of-box), 2) fine-tuned on the original training data (org), 3) fine-tuned on revision version 1 of the training data (r1), and 4) fine-tuned on revision version 2 of the training data (r2). To clearly illustrate the impact of revised training data, we include the score differences compared to the original dataset (org) for out-of-box, r1 and r2 ($\Delta$ over org.).

\begin{table*}[t]
 \caption{Summarization performance across different training data sets and student LLMs. Mean of auto-evaluator rating (1-5 scale) is given. Superscript * marks performance better than teacher LLM (GPT-3.5). Best scores for each channel are bold-faced. r: mean of auto-rating, $\Delta$: difference over when trained on the original training dataset}
  \centering
  \small
  \begin{tabular}{lllllllll}
      \toprule
&\multicolumn{2}{c}{\textbf{Bot Chat}}&\multicolumn{2}{c}{\textbf{Teammate Chat}} & \multicolumn{2}{c}{\textbf{WebForm}} & \multicolumn{2}{c}{\textbf{Email}}\\
&\textbf{r}&\textbf{$\Delta$} &r&$\Delta$ &r&$\Delta$&r&$\Delta$\\
 \midrule
GPT3.5 (out-of-box)&$2.525$&&$3.965$&&$4.030$&&$1.485$&\\
\midrule
Llama 3.1 8B (out-of-box)&$1.630$&$-0.850$&$3.975^{*}$&$0.030$&$3.625$&$-0.565$&$1.185$&$-0.400$\\
Llama 3.1 8B (org)&$2.480$&&$3.945$&&$4.190^{*}$&&$1.585^{*}$&\\
Llama 3.1 8B (r1)&$3.435^{*}$&$0.955$&$4.370^{*}$&$0.425$&$\mathbf{4.325^{*}}$&$0.135$&$2.060^{*}$&$0.475$\\
Llama 3.1 8B (r2)&$3.265^{*}$&$0.785$&$4.440^{*}$&$0.495$&$4.005$&$-0.185$&$\mathbf{2.062^{*}}$&$0.477$\\
\midrule
Llama 3.2 3B (out-of-box)&$1.195$&$-0.595$&$2.470$&$-0.560$&$2.425$&$-0.580$&$1.182$&$-0.134$\\
Llama 3.2 3B (org)&$1.790$&&$3.030$&&$3.005$&&$1.316$&\\
Llama 3.2 3B (r1)&$3.200^{*}$&$1.410$&$4.260^{*}$&$1.230$&$3.885$&$0.880$&$1.910^{*}$&$0.594$\\
Llama 3.2 3B (r2)&$3.220^{*}$&$1.430$&$\mathbf{4.470^{*}}$&$1.440$&$3.875$&$0.870$&$1.721^{*}$&$0.405$\\
\midrule
Llama 3.2 1B (out-of-box)&$1.375$&$-0.815$&$1.400$&$-2.075$&$1.625$&$-2.155$&$1.000$&$-0.721$\\
Llama 3.2 1B (org)&$2.190$&&$3.475$&&$3.780$&&$1.336$&\\
Llama 3.2 1B (r1)&$3.030^{*}$&$0.840$&$4.040^{*}$&$0.565$&$3.905$&$0.125$&$1.600^{*}$&$0.264$\\
Llama 3.2 1B (r2)&$3.165^{*}$&$0.975$&$4.035^{*}$&$0.560$&$3.720$&$-0.060$&$1.689^{*}$&$0.353$\\
\midrule
Qwen3 4B (out-of-box)&$2.345$&$0.250$&$4.080^{*}$&$0.395$&$3.975$&$-0.035$&$1.594^{*}$&$0.094$\\
Qwen3 4B (org)&$2.095$&&$3.685$&&$4.010$&&$1.500^{*}$&\\
Qwen3 4B (r1)&$3.000^{*}$&$0.905$&$4.150^{*}$&$0.465$&$4.115^{*}$&$0.105$&$1.633^{*}$&$0.133$\\
Qwen3 4B (r2)&$3.065^{*}$&$0.970$&$4.350^{*}$&$0.665$&$3.835$&$-0.175$&$1.535^{*}$&$0.035$\\
\midrule
Gemma3 1B (out-of-box)&$1.265$&$-0.700$&$2.290$&$-1.015$&$1.320$&$-2.460$&$1.022$&$-0.016$\\
Gemma3 1B (org)&$1.965$&&$3.305$&&$3.780$&&$1.038$&\\
Gemma3 1B (r1)&$\mathbf{3.520^{*}}$&$1.555$&$3.235$&$-0.070$&$3.000$&$-0.780$&$1.027$&$-0.011$\\
Gemma3 1B (r2)&$2.065$&$0.100$&$3.400$&$0.095$&$3.350$&$-0.430$&$1.130$&$0.092$\\
\midrule
Gemma3 270M (out-of-box)&$1.050$&$-0.880$&$1.170$&$-1.415$&$1.180$&$-1.875$&$1.014$&$-0.008$\\
Gemma3 270M (org)&$1.930$&&$2.585$&&$3.055$&&$1.022$&\\
Gemma3 270M (r1)&$2.120$&$0.190$&$2.645$&$0.060$&$2.675$&$-0.380$&$1.049$&$0.027$\\
Gemma3 270M (r2)&$2.130$&$0.200$&$2.690$&$0.105$&$2.895$&$-0.160$&$1.022$&$0.000$\\
\midrule
Phi3.5mini 4B (out-of-box)&$1.610$&$-0.305$&$3.645$&$-0.160$&$3.080$&$-0.890$&$1.535^{*}$&$0.067$\\
Phi3.5mini 4B (org)&$1.915$&&$3.805$&&$3.970$&&$1.468$&\\
Phi3.5mini 4B (r1)&$2.285$&$0.370$&$4.435^{*}$&$0.630$&$3.960$&$-0.010$&$1.675^{*}$&$0.207$\\
Phi3.5mini 4B (r2)&$2.290$&$0.375$&$4.305^{*}$&$0.500$&$3.775$&$-0.195$&$1.792^{*}$&$0.324$ \\
\midrule
  \end{tabular}
  \label{tab:perf_on_training_set_llm}
\end{table*}

Table~\ref{tab:perf_on_training_set_llm} shows that revising the teacher-generated training set yields broad and substantial gains in student fine-tuning performance across channels. Relative to fine-tuning on the original dataset (“org”), training on the revised datasets improves BotChat performance for all student models (7/7): r1 increases the mean rating by +0.89 points on average (range +0.19 to +1.56), and r2 by +0.69 points (range +0.10 to +1.43). The effect is similarly consistent on TeammateChat, where r1 improves 6/7 models (mean +0.47) and r2 improves 7/7 (mean +0.55, reaching gains up to +1.44 for Llama 3.2 3B). Email summarization also benefits: r1 improves 6/7 models (mean +0.24) and r2 improves 6/7 (with 1/7 unchanged; mean +0.24). Overall, these results shows that targeted revisions to the distillation/fine-tuning data are a primary driver of downstream student quality, beyond simply performing fine-tuning on the original teacher outputs.

The revised datasets also enable student models to match or surpass the teacher more reliably across the channels. Under r1, both Llama 3.1 8B and Qwen3 4B outperform GPT-3.5 in all four channels simultaneously (e.g., Llama 3.1 8B r1: 3.435 vs 2.525 on BotChat, 4.370 vs 3.965 on TeammateChat, 4.325 vs 4.030 on WebForm, and 2.060 vs 1.485 on Email). Consistent with this, the best student score in each channel is achieved by a revised-data variant rather than “org” or out-of-box: Gemma3 1B (r1) is best on BotChat (3.520), Llama 3.2 3B (r2) is best on Teammate Chat (4.470), Llama 3.1 8B (r1) is best on WebForm (4.325), and Llama 3.1 8B (r2) is best on Email (2.062). This pattern indicates that dataset revision is not merely stabilizing performance, but is instrumental in reaching peak student capability.

Revisions are particularly valuable in cases where naive fine-tuning on “org” is counterproductive. For example, Qwen3 4B degrades from out-of-box to org on both BotChat (2.345 → 2.095) and Teammate Chat (4.080 → 3.685), but revised-data training recovers and substantially exceeds both baselines (Bot Chat 3.000–3.065, Teammate Chat 4.150–4.350). A similar recovery is visible for Phi3.5mini 4B on Email (org 1.468 vs out-of-box 1.535, rising to 1.675–1.792 with r1/r2).

\subsection{Revision quality matters.}

Although revised training data generally improves students, Table~\ref{tab:perf_on_training_set_llm} indicates that downstream gains depend on the \emph{quality} and \emph{channel-compatibility} of each revision step. This is most visible for WebForm when comparing r1 and r2. Relative to org, r1 yields near-neutral WebForm changes overall (improving 4/7 student models; mean $\Delta \approx +0.01$), whereas r2 improves only 1/7 models and exhibits a negative mean change ($\Delta \approx -0.05$). The degradation is consistent across families, e.g., Llama 3.1 8B drops from $4.190$ (org) to $4.005$ (r2; $\Delta=-0.185$), Qwen3 4B from $4.010$ to $3.835$ ($\Delta=-0.175$), and Phi3.5mini 4B from $3.970$ to $3.775$ ($\Delta=-0.195$), suggesting that the additional revision step introduced in r2 is not uniformly beneficial for this channel.

Table~\ref{tab:revision_success_rate} helps explain this pattern by quantifying \emph{revision success rate}, where a revision is counted as successful if it fixes the targeted error or leaves the summary unchanged (thus preserving or improving quality), and as a failure if it degrades summary quality. The WebForm-specific correction used to generate r1, \texttt{unn\_content\_webform\_email\_copy}, is highly reliable (97\% success). In contrast, the additional r2 step---a global redundancy-removal correction (\texttt{unn\_content\_redundant}) applied to all channels as the final prompt in the cascade---has substantially lower success on WebForm (61\%), implying that a non-trivial fraction of WebForm training labels are worsened by this step. This failure rate is also channel-dependent: the same redundancy correction is much more reliable for TeammateChat (97\%) and Email (85\%), which is consistent with r2 being beneficial or neutral outside WebForm.

\begin{table*}
 \caption{Summary revision success rate per error type}
  \centering
  \small
  \begin{tabular}{llllll}
    \toprule
    \textbf{Error type}& \textbf{Bot Chat} & \textbf{WebForm} & \textbf{TeammateChat} & \textbf{Email} & \textbf{Overall}  \\
    \midrule
    unn\_content\_requests\_agent & 94\% & - & - & - & 94\% \\
    sentiment\_inferred\_frustrated & 92\% & - & - & - & 92\% \\
    unn\_content\_webform\_email\_copy & - & 97\% & - & - & 97\% \\
    unn\_content\_transfer & - & - & 99\% & - & 99\% \\
    unn\_content\_courtesy  & - & - & 98\% & 86\% & 92\% \\
    content\_order & - & - & - & 84\% & 84\% \\
    unn\_content\_email\_footer & - & - & - & 91\% & 91\% \\
    unn\_content\_redundant & 78\% & 61\% & 97\% & 85\% & 80\% \\
    \bottomrule
  \end{tabular}
  \label{tab:revision_success_rate}
\end{table*}

A second signal that revision effects are not uniform across students is Gemma3 1B: it achieves the best BotChat score with r1 ($3.520$) yet drops sharply on WebForm from $3.780$ (org) to $3.000$ (r1; $\Delta=-0.780$). Since this regression occurs prior to the r2 redundancy step, it suggests negative transfer under multi-channel fine-tuning. Many r1 edits are deletion-style and highly successful in Table~\ref{tab:revision_success_rate} (e.g., 92--99\% for several chat-specific errors), which can shift the target distribution toward more aggressively compressed summaries. Lower-capacity students may over-generalize this behavior across channels, harming WebForm where completeness of structured details is critical. 

Overall, these results indicate that revision quality is a key determinant of ARF performance. In particular, revisions that are reliable in one channel (or on one error type) may be substantially riskier in another, and globally applied corrections such as redundancy removal should be treated as channel-dependent operations that may require tighter constraints or validation for structured inputs such as WebForms.

\subsection{The better the revision, the better the fine-tuning}

To quantify how revision quality translates into downstream student performance, we conducted a manual assessment of the \emph{overall summary quality change} introduced by our error-correction cascade for both r1 and r2. For each intermediate output in the cascade, we compared the revised summary against the pre-prompt summary as a baseline and assigned a discrete quality-change score: a successful improvement that corrected an error was scored as $+1$, the introduction of a new error or degradation in overall quality was scored as $-1$, and revisions producing similar-quality summaries were scored as $0$. We then computed the \emph{cumulative quality change} for each sample by summing these scores across all prompts applied for that revision version. This procedure was performed on 100 samples per channel for r1 and r2, and Table~\ref{tab:cumulative_quality_change} reports the mean cumulative quality change.

Table~\ref{tab:cumulative_quality_change} shows that revision quality varies substantially across channels and between r1 and r2. BotChat exhibits a strong positive cumulative change under r1 ($0.65$), which slightly decreases under r2 ($0.59$). TeammateChat shows the largest quality gains overall, increasing from $0.58$ in r1 to $1.26$ in r2. In contrast, WebForm has a small positive cumulative change that remains unchanged between r1 and r2 ($0.12$), while Email drops sharply from $0.41$ (r1) to $0.14$ (r2). 

These channel-level differences are consistent with the per-error revision reliability reported in Table~\ref{tab:revision_success_rate}. 
Redundancy removal (\texttt{unn\_content\_redundant}), 
the final prompt distinguishing r2 from r1, is highly reliable for TeammateChat (97\% success), which aligns with the large increase in cumulative quality change for TeammateChat when moving from r1 to r2. In contrast, redundancy removal is substantially less reliable for BotChat (78\%) and especially for WebForm (61\%), providing a plausible explanation for why r2 yields smaller net quality improvements in BotChat and does not increase WebForm quality beyond r1. Email exhibits moderate redundancy-removal success (85\%), yet the cumulative gain decreases markedly in r2, suggesting that even modest failure rates can dominate when the correction is applied broadly and interacts with other channel-specific edits.

\begin{table}[h]
\centering
\caption{Cumulative summary quality change for r1/r2 from manual review}
\label{tab:overall_summary_quality_manual}
\small
\begin{tabular}{lcccc}
\toprule
 & \textbf{BotChat} & \textbf{TeammateChat} & \textbf{WebForm} & \textbf{Email} \\
\midrule
\textbf{r1} & 0.65 & 0.58 & 0.12 & 0.41 \\
\textbf{r2} & 0.59 & 1.26 & 0.12 & 0.14 \\
\bottomrule
\label{tab:cumulative_quality_change}
\end{tabular}
\end{table}

\begin{figure}[t] 
    \centering
    \includegraphics[width=0.53\textwidth]{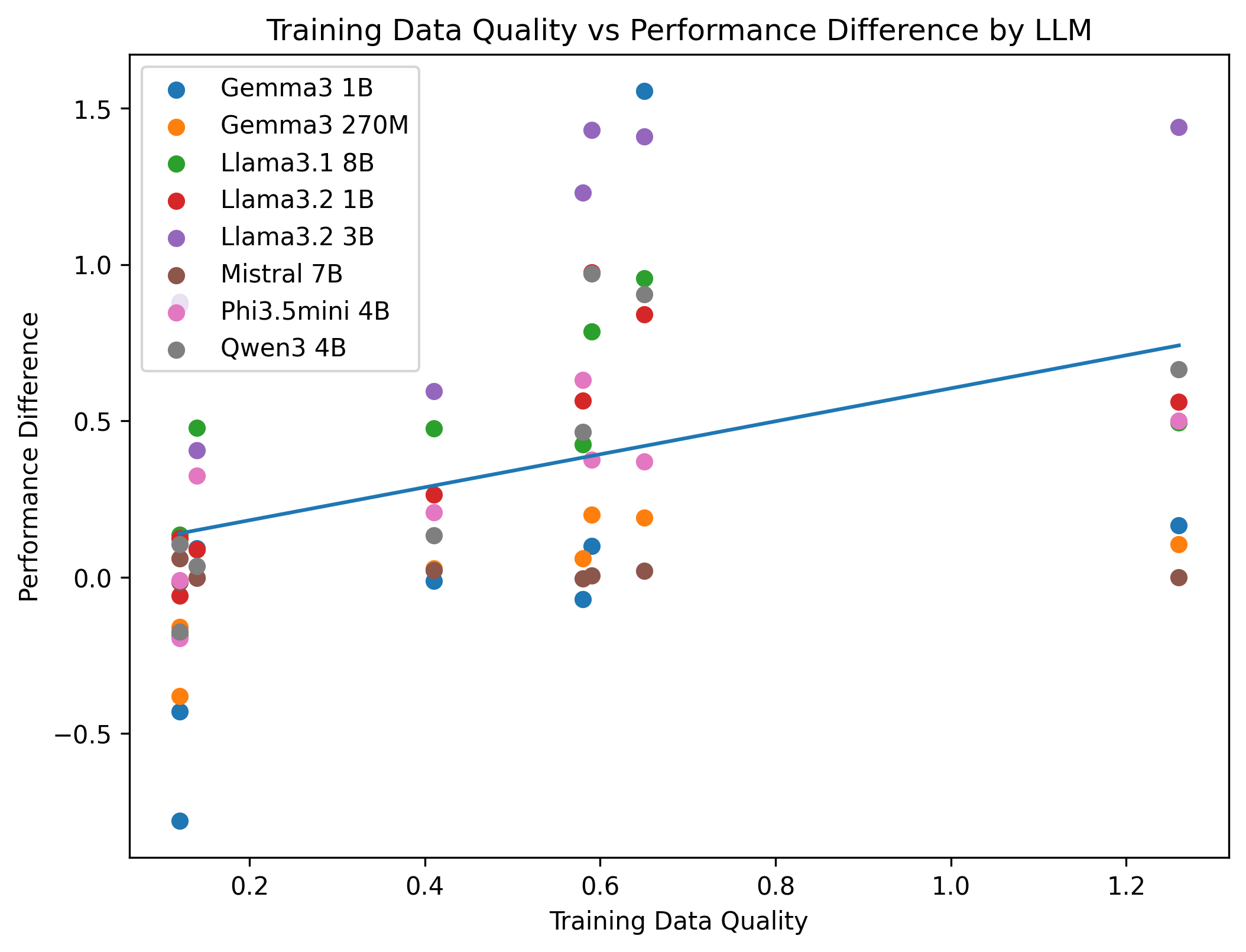}
    \caption{Correlation between training data quality and performance improvement}
    \label{fig:train_quality_vs_delta}
\end{figure}

Finally, we test whether this manual measure of revision quality predicts fine-tuning effectiveness. Using the cumulative quality change as a proxy for training-label quality, we performed correlation analysis against fine-tuning gains measured as $\Delta$ over org across all student models, channels, and revision versions (Figure~\ref{fig:train_quality_vs_delta}). The results show a consistent positive association between revision quality and downstream improvement: Pearson correlation is $0.44$, Spearman rank correlation is $0.60$, and Kendall's $\tau$ is $0.46$. The stronger rank-based correlations suggest that the relationship is monotonic but not strictly linear, which is plausible given model heterogeneity (capacity/architecture) and the fact that revision quality values are channel/version-specific aggregates.
Collectively, these findings motivate investing in more reliable revision prompts and channel-aware validation, as gains in revision quality translate into measurable gains in student performance.



\section{Discussion}
We show that compact open-source LLMs can exceed GPT-3.5 on customer-service interaction summarization when trained with our Analyze--Revise--Finetune pipeline. In particular, Llama 3.1 8B and QWen3 4B outperform GPT-3.5 across BotChat, TeammateChat, Email, and WebForm, demonstrating that specialized production workloads can be served by smaller models with lower cost and reduced exposure of sensitive data. OpenAI does not disclose GPT-3.5's parameter count; using GPT-3's reported 175B parameters~\cite{brown2020nips} and secondary estimates suggesting a similar scale for GPT-3.5~\cite{10.1371/journal.pone.0288453, luo2024moelora}, we treat our 270M--8B student models as orders of magnitude smaller than the teacher.

Our error-focused evaluation scheme ensures a consistent and systematic evaluation of summary quality, allowing for precise identification of the most common errors requiring correction. Additionally, by separating the editor LLM from the teacher LLM and utilizing Llama 3.1 70B for training data revisions, our approach further optimizes both inference cost and data privacy, offering substantial advantages compared to approaches that directly employ large teacher models for revisions.

Although performance gains were observed across different student models using the revised training datasets, the quality of the revisions emerged as an important factor influencing final model performance. Thus, future work should focus on developing more robust and reliable revision processes to further improve the overall effectiveness and dependability of the training data revision step.


\section{Conclusion}
We propose Analyze--Revise--Finetune, a targeted error-correction fine-tuning pipeline that improves training data derived from a proprietary teacher to train smaller open-source students. On a real production summarization system spanning four channels, the resulting student models match or surpass GPT-3.5 while improving cost efficiency and privacy. The ARF pipeline can be successfully extended to a wide range of LLM-based applications, offering consistent improvements in cost efficiency, output quality, and data privacy.

\section*{Acknowledgments}
We'd like to thank Mehran Nadjarbashi Noghani, Swapnil Thorat, Ryan Wen, Zonghan Xu, Pooja Daryani, and Megha Korade for their support in this work. 

\bibliographystyle{unsrt}  
\bibliography{references}

\appendix

\section{Additional Table}

\begin{table}[h]
 \caption{Summary error types as defined in the summary evaluation scheme}
  \centering
{\footnotesize
\begin{tabular}{llp{7.8cm}}
\hline
\textbf{Primary Label} & \textbf{Sub Label} & \textbf{Description} \\
\hline
\multirow{5}{*}{Content}&content\_missing&\multirow{4}{7.5cm}{\parbox{7.5cm}{Error in summary content - 1) missing critical content, 2) not following a logical chronological order, 3) inaccurate representation of original content, or 4) inclusion of completely fabricated content} }\\
&content\_order&\\
&content\_inaccurate&\\
&content\_hallucination&\\
\cmidrule(r){2-3}
&content\_other&Any other content issues \\
\hline

\multirow{19}{*}{Entities}&entity\_missing\_item\_number&\multirow{8}{7.5cm}{\parbox{7.5cm}{An entity (i.e., item number, item name, order number, return number, case number, username, price, or transaction ID) that is represented in the original content is missing in summary}}\\
&entity\_missing\_item\_name&\\
&entity\_missing\_order\_number&\\
&entity\_missing\_return\_number&\\
&entity\_missing\_case\_number&\\
&entity\_missing\_username&\\
&entity\_missing\_price\\
&entity\_missing\_transaction\_id&\\
\cmidrule(r){2-3}
&entity\_inaccurate\_item\_number&\multirow{8}{7.5cm}{\parbox{7.5cm}{An entity (i.e., item number, item name, order number, return number, case number, username, price, or transaction ID) that is represented in the original content is inaccurately presented in summary}}\\
&entity\_inaccurate\_item\_name&\\
&entity\_inaccurate\_order\_number&\\
&entity\_inaccurate\_return\_number&\\
&entity\_inaccurate\_case\_number&\\
&entity\_inaccurate\_username&\\
&entity\_inaccurate\_price&\\
&entity\_inaccurate\_transaction\_id&\\
\cmidrule(r){2-3}
&entity\_inaccurate\_other&When inaccurate entity type is unclear due to redaction\\
&entity\_other&Any other entity-related issues\\
\hline

\multirow{3}{*}{Data Elements}&data\_element\_missing&\multirow{2}{7.5cm}{\parbox{7.5cm}{A data element (e.g., dates) is missing or inaccurate in summary}}\\
&data\_element\_inaccurate&\\
\cmidrule(r){2-3}
&data\_element\_other&Any other issues related to data elements \\
\hline

\multirow{3}{*}{Customer Type}&customer\_type\_missing&\multirow{2}{7.5cm}{\parbox{7.5cm}{Customer type is missing or inaccurate in summary. Applicable only to WebForm.}}\\
&customer\_type\_inaccurate&\\
\cmidrule(r){2-3}
&customer\_type\_other&Other Issues related to customer type \\
\hline

\multirow{9}{1.8cm}{\parbox{1.8cm}{Unnecessary information}}&unn\_content\_redundant&\multirow{8}{7.5cm}{\parbox{7.5cm}{Summary includes unnecesary information, i.e., 1) redundant or repetitive information, 2) courtesy or plesantry, 3) bot's response, 4) customer ranting, 5) lack of specific details in the original content, 6) requets for human agent, 7) transfer to next agent, 8) email footer/signature, 9) request for email copy of original webform}}\\
&unn\_content\_courtesy&\\
&unn\_content\_platform\_response\_included&\\
&unn\_content\_customer\_rant&\\
&unn\_content\_no\_details&\\
&unn\_content\_requests\_agent&\\
&unn\_content\_transfer&\\
&unn\_content\_email\_footer&\\
&unn\_content\_webform\_email\_copy&\\
\cmidrule(r){2-3}
&unn\_other&Any other type of unnecessary information \\
\hline

\multirow{6}{1.8cm}{\parbox{1.8cm}{Inferred Sentiment}}&sentiment\_inferred\_confused&\multirow{2}{7.5cm}{\parbox{7.5cm}{Inferred customer sentiment, i.e., confusion or frustration, not in the original content but in summary}}\\
&sentiment\_inferred\_frustrated&\\
\cmidrule(r){2-3}
&sentiment\_inferred\_not\_confused&\multirow{3}{7.5cm}{\parbox{7.5cm}{Inferred info. of customer not showing sentiment, i.e., confusion, frustration, or complaint, not in the original content but in summary}}\\
&sentiment\_inferred\_not\_frustrated&\\
&sentiment\_inferred\_no\_complaint&\\
\cmidrule(r){2-3}
&sentiment\_other&Any other sentiment-related issues\\
\hline

\multirow{3}{*}{Language}&language\_non\_english\_not\_identified&Missing the fact that oroginal content is in a non-English\\
&language\_translation\_inaccurate&Inaccuracy in summary due to translation\\
&language\_other&Any other issues related to language\\
\hline
\end{tabular}
}
\label{table:eval_criteria_error_types}
\end{table}

\section{Additional Figures}

\begin{figure*}[h] 
\centering
\tiny
\begin{tcolorbox}[colback=gray!10, colframe=black, title={Prompt to remove \texttt{unn\_content\_requests\_agent} error from BotChat summaries} ]
Read the following summary of a chat transcript between a customer and an automatic service agent.
Rewrite the summary to remove any information about the customer wanting or requesting to connect with a human agent.
If there's no other content in the summary other than customer wanting or requesting to connect with a human agent, say "<ul><li>nothing to summarize</li></ul>".
Keep the original format of the summary that is unordered list items in HTML.
Do not add anything else other than the revised summary.
\end{tcolorbox} 

\begin{tcolorbox}[colback=gray!10, colframe=black, title={Prompt to remove \texttt{sentiment\_inferred\_frustrated} error from BotChat summaries} ]
Read the following summary of a chat transcript between a customer and an automatic service agent.
Rewrite the summary to remove any information about the customer's emotion such as being frustrated, any information about the reason of the emotion, and any information about how the customer expressed their emotion.
If there's no other content in the summary other than customer's emotion, say "<ul><li>nothing to summarize</li></ul>".
Keep the original format of the summary that is unordered list items in HTML.
Do not add anything else other than the revised summary.
\end{tcolorbox} 

\begin{tcolorbox}[colback=gray!10, colframe=black, title={Prompt to remove \texttt{unn\_content\_courtesy} error from TeammateChat summaries} ]
Read the following summary.
Rewrite the summary to remove any information regarding courteous engagement between agent and customer. 
Courteous engagement includes expressing gratitude, assuring customer for resolution of their issue, affirming customer's loyalty of having acocunt for many years, among others.
Do not remove information regarding reassuring or confirming action if the action is conveying additional details of customer's issue or agent's resolution such as eBay Money Back Guarantee or timeframe.
Do not remove any important information such as Item Ids, Return Ids, Case Ids, etc or customer's role as a buyer or a seller.
Keep the original format of the summary that is unordered list items in HTML.
Do not add anything else other than the revised summary.
\end{tcolorbox} 

\begin{tcolorbox}[colback=gray!10, colframe=black, title={Prompt to remove \texttt{unn\_content\_transfer} error from TeammateChat summaries} ]
Read the following summary.
Rewrite the summary to remove any information regarding customer being transferred to a specialist or to another agent.
Do not remove any important information such as Item Ids, Return Ids, Case Ids, etc or customer's role as a buyer or a seller.
Keep the original format of the summary that is unordered list items in HTML.
Do not add anything else other than the revised summary.
\end{tcolorbox} 

\begin{tcolorbox}[colback=gray!10, colframe=black, title={Prompt to remove \texttt{unn\_content\_webform\_email\_copy} error from WebForm summaries} ]
Read the following summary of a WebForm submitted by a customer.
Rewrite the summary to remove any information about the customer wanting or requesting a copy of the communication to their email.
Do not add anything else other than the revised summary.
\end{tcolorbox} 

\begin{tcolorbox}[colback=gray!10, colframe=black, title={Prompt to remove \texttt{unn\_content\_redundant} for both BotChat, TeammateChat, and WebForm summaries} ]
Read the following summary.
Rewrite the summary to remove any information that is repeatedly stated. Do not remove any important information such as Item Ids, Return Ids, Case Ids, etc or customer's role as a buyer or a seller.
Keep the original format of the summary that is unordered list items in HTML.
Do not add anything else other than the revised summary.
\end{tcolorbox}

\begin{tcolorbox}[colback=gray!10, colframe=black, title={Prompt to remove \texttt{content\_order} for Email summaries} ]
Read the following email exchanges between an eBay customer service agent and an eBay customer, along with its summary.

Reorganize the summary according to the following principles:

Chronological Order: Arrange events in the order they occurred—older events should appear first, followed by newer events.
General to Specific: High-level descriptions of events should come first, followed by more detailed descriptions.

Instructions:
Rearrange the order of bullet points only. Do not rewrite or remove any bullet point.
Keep the original format of the summary that is unordered list items in HTML.
Do not add anything else other than the revised summary.
\end{tcolorbox}

\begin{tcolorbox}[colback=gray!10, colframe=black, title={Prompt to remove \texttt{unn\_content\_courtesy} for Email summaries} ]
Read the following summary.
Rewrite the summary to remove any information regarding courteous engagement between agent and customer. 
Courteous engagement includes expressing gratitude, assuring customer for resolution of their issue, affirming customer's royalty of having account for many years, among others.
Do not remove information regarding reassuring or confirming action if the action is conveying additional details of customer's issue or agent's resolution such as eBay Money Back Guarantee or timeframe.
Do not remove any important information such as Item Ids, Return Ids, Case Ids, etc or customer's role as a buyer or a seller.
Keep the original format of the summary that is unordered list items in HTML.
Do not add anything else other than the revised summary.
\end{tcolorbox} 

\begin{tcolorbox}[colback=gray!10, colframe=black, title={Prompt to remove \texttt{unn\_content\_email\_footer}  for Email summaries} ]
Read the following summary.
Rewrite the summary to remove any information regarding email footers or signature. 
Email footers or signatures includes customer's location.
Also remove any bullet point stating "User wants a copy of the communication sent to their email address.".
Do not remove any important information such as Item Ids, Return Ids, Case Ids, etc or customer's role as a buyer or a seller.
Keep the original format of the summary that is unordered list items in HTML.
Do not add anything else other than the revised summary.
\end{tcolorbox} 

\begin{tcolorbox}[colback=gray!10, colframe=black, title={Prompt to remove \texttt{unn\_content\_redundant} for Email summaries} ]
Read the following summary. Rewrite the summary to remove any information repeatedly stated.
Do not remove any details even when there's a general version of the same information. Instead remove general version.
When the same information is stated more than twice, keep only one description.
Do not remove essential details such as Item IDs, Return IDs, Case IDs, Ticket Numbers, Dates, or the customer’s role (buyer or seller).
Keep the original format of the summary that is unordered list items in HTML.
Do not add anything else other than the revised summary.
\end{tcolorbox}

\caption{Prompts for summary error correction}
\label{fig:error_correction_prompts}
\end{figure*}

\end{document}